\documentclass[11pt,letterpapers]{article}

\usepackage{booktabs}
\usepackage{url}
\usepackage{epsfig}
\usepackage{xcolor}
\usepackage{authblk}
\usepackage{hyperref}       
\usepackage{amsfonts}       
\usepackage{nicefrac}       
\usepackage{microtype}      
\usepackage{xcolor}         
\usepackage{amsmath}

\usepackage[top=1in,bottom=1in,left=1in,right=1in,marginparwidth=1in]{geometry}
\usepackage[margin=0.5in,font=small,labelfont=bf]{caption}

\usepackage{scrextend}
\deffootnote[0.55in]{0pt}{1.8em}{\makebox[1.3em][l]{\thefootnotemark}}

\usepackage{graphicx}
\graphicspath{{figures/}}

\usepackage{pifont}
\newcommand{\cmark}{\ding{51}}%
\newcommand{\xmark}{\ding{55}}%

\newcommand{\data}{PGB}

\newcommand{\neurips}{OGB}
\newcommand{\kdd}{HGB}
\newcommand{\storc}{S2ORC}

\title{\data: A PubMed Graph Benchmark for Heterogeneous Network Representation Learning}

\author[ ]{Eric W Lee}
\author[ ]{Joyce C Ho}
\affil[ ]{Department of Computer Science, Emory University}
\affil[ ]{\textit{\{ewlee4, joyce.c.ho\}@emory.edu}}
\date{}

\begin{document}

\maketitle

\abstract{
There has been rapid growth in biomedical literature, yet capturing the heterogeneity of the bibliographic information of these articles remains relatively understudied.
Although graph mining research via heterogeneous graph neural networks has taken center stage, it remains unclear whether these approaches capture the heterogeneity of the PubMed database, a vast digital repository containing over 33 million articles.  
We introduce PubMed Graph Benchmark (PGB), a new benchmark dataset for evaluating heterogeneous graph embeddings for biomedical literature.
The benchmark contains rich metadata including abstract, authors, citations, MeSH terms, MeSH hierarchy, and some other information.
The benchmark contains three different evaluation tasks encompassing systematic reviews, node classification, and node clustering.
In PGB, we aggregate the metadata associated with the biomedical articles from PubMed into a unified source and make the benchmark publicly available for any future works.
}
\section{Introduction}

Academic graphs generated from bibliographic data serve as an essential data source across many different fields.
Analysis of such graphs can be used for personalized article recommendations \cite{west2016recommendation}, retrieval of relevant articles \cite{beel2016paper}, understanding of trends in the field \cite{asatani2018detecting,effendy2017analysing}, measuring academic influence and novelty \cite{zhang2021measuring}, and identifying relevant academic communities \cite{chakraborty2013overcite}.
PubMed is an example of an academic graph that contains over 33 million citations and abstracts of literature related to biomedicine and health fields as well as related disciplines such as life sciences, behavioral sciences, chemical sciences, and bioengineering \cite{pubmed}.
PubMed articles have been used to perform numerous systematic reviews (SR) \cite{cohen2006reducing,jain2018learning,sys_review}, evaluate biological processes \cite{bioprocess}, identify protein-protein interactions \cite{proteinabstract}, curate genes \cite{curategene}, and extract biological networks \cite{yang2020heterogeneous}.
To date, much of the work on PubMed literature has focused on mining the text.
However, the rich citation structure can be utilized to automate the SR process and provide better representation than their textual counterparts \cite{lee2020mmidas}.

For analysis of the academic graphs, low-dimensional representations, or embeddings, of the graph's nodes, serve as the fundamental analysis tool \cite{cai2018comprehensive, tsitsulin2021frede, yang2020scaling}.
The idea is to learn a compact representation of each node that preserves the structural information and properties of the graph.
The graph embedding can then be used for a variety of downstream tasks such as node retrieval/recommendation \cite{zhou2017scalable}, node classification \cite{wang2017community}, node clustering \cite{nie2017unsupervised}, and link prediction \cite{wei2017cross}.
In recent years, graph neural networks (GNNs) \cite{battaglia2018relational,gori2005new,kipf2016semi,li2015gated,scarselli2008graph} have become pervasive due to their impressive performances across various tasks.
However, GNNs have been predominantly studied in the homogeneous network setting, where there is only a single node type and link-type \cite{tsitsulin2021frede,yang2020homogeneous,yang2020scaling}.
Yet, an academic graph can contain multiple objects (nodes) and link types (edges)  including author information, venue information, and keywords.
As such, researchers are exploring extending GNNs to the heterogeneous information network (HIN) with multiple node types and edges, each with potentially different side information.
Heterogeneous GNNs have been proposed to incorporate the node and edge types \cite{kipf2016semi, wang2019heterogeneous, yang2020heterogeneous, yang2020interpretable, yun2019graph}.

Despite the development of such models, a recent paper demonstrated that in fact, the results generated by these state-of-the-art heterogeneous GNNs were merely a mirage \cite{lv2021we}.
The lack of consistent experimental setup and preprocessing of the data led to widely varying results. In some cases, vanilla GNN models were better than their heterogeneous GNN counterparts.
As such, there have been recent developments toward developing benchmark graph datasets.
\neurips~was developed as a large-scale benchmark for a broad range of graph machine learning tasks \cite{hu2020open}.
It encompasses various domains, including bibliographic data (i.e., arXiv and Microsoft Academic Graph (MAG) \cite{wang2020microsoft}). The ogbn-arxiv and obgn-papers100M datasets focus on the simple citation network using a homogeneous network representation whereas the obgn-mag extracts a heterogeneous information network from MAG and contains four different node types (papers, authors, institutions, and topics) along with their links. However, \neurips~is more geared toward evaluating homogeneous GNNs rather than the heterogeneous GNNs.

\kdd~\cite{lv2021we} was developed as a new medium-scale benchmark dataset that spanned 11 heterogeneous networks including two bibliographic datasets: (1) DBLP, a citation network of computer science that contains four nodes (authors, papers, terms, and venues) and (2) ACM, a citation network that spans papers from 5 conference venues that contains three nodes (authors, papers, and subjects). \kdd~also contains a PubMed-based dataset involving a network of genes, diseases, chemicals, and species extracted from the articles using Named Entity Recognition software.
However, the experimental studies of GNN models on \kdd~demonstrated substantially better performance on DBLP and ACM datasets than on the PubMed dataset.
This may not be surprising as many general domain text mining models fail to generalize to biomedical literature \cite{bannach2019machine,cohen2008optimizing,cohen2006reducing}, motivating a more extensive study.

\begin{figure*}[t!]
    \centering
    \begin{minipage}{0.45\linewidth}
        \center
            \includegraphics[width=0.85\textwidth]{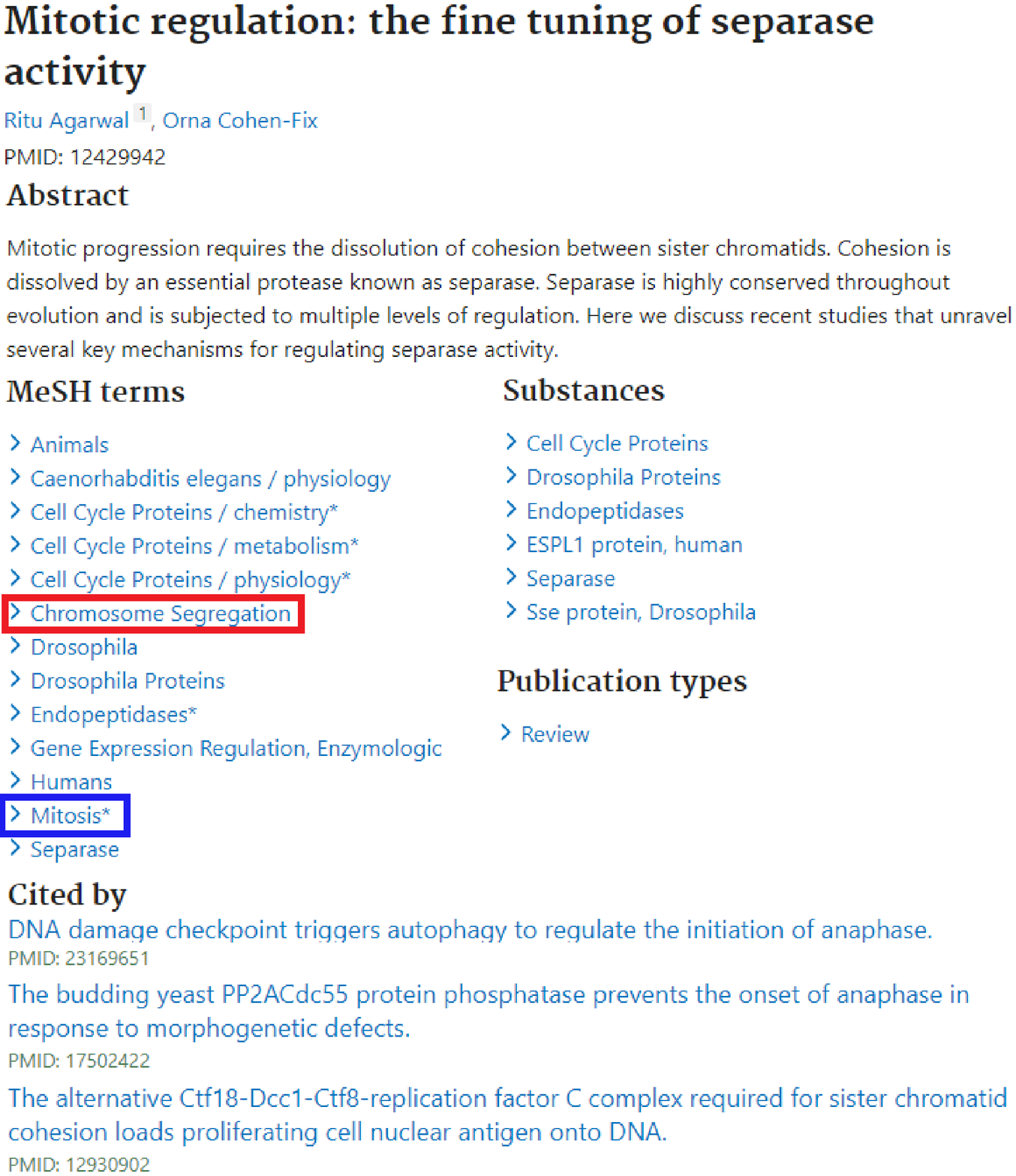}
    \end{minipage}
    \begin{minipage}{0.45\linewidth}
        \center
            \includegraphics[width=0.9\textwidth]{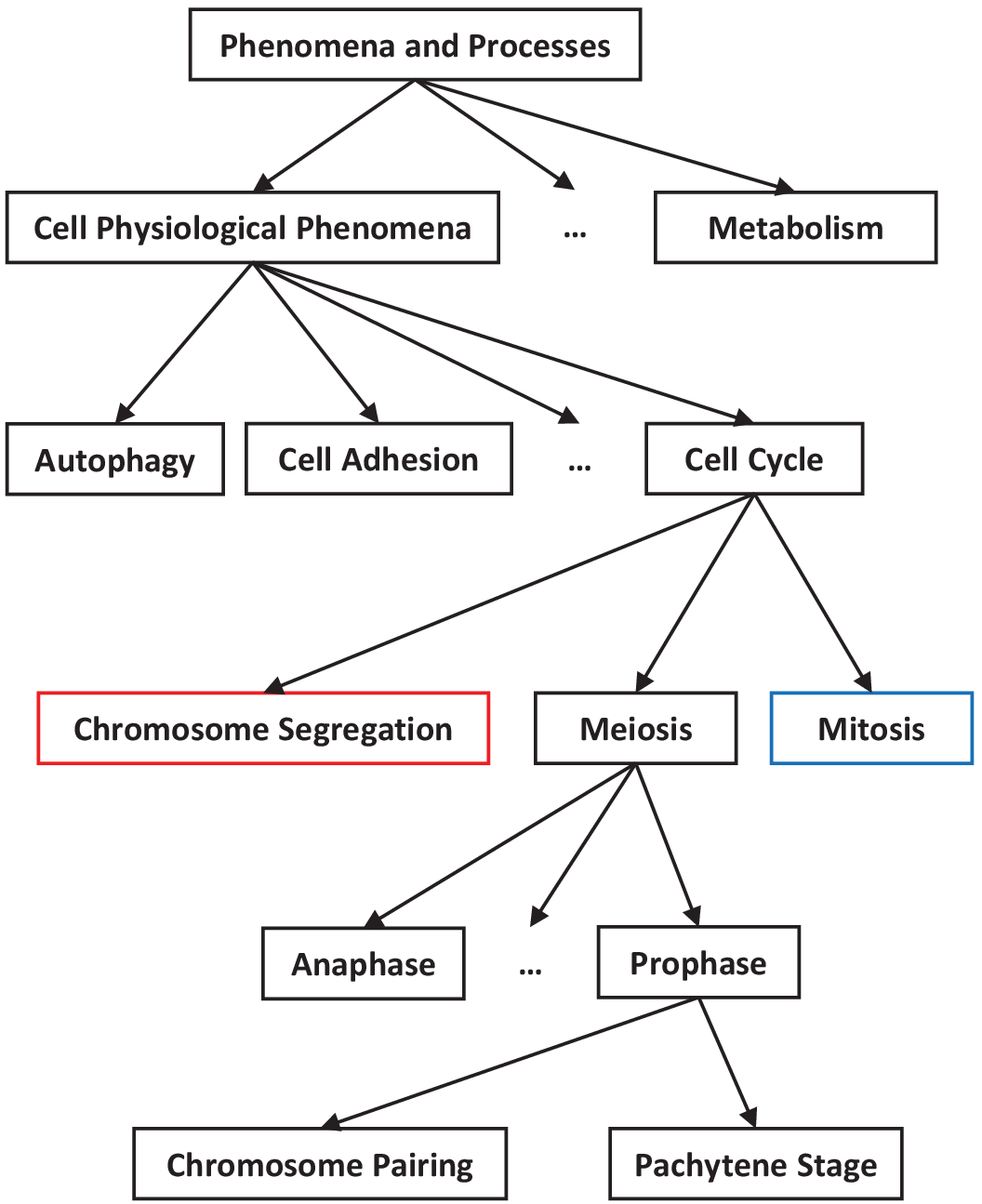}
    \end{minipage}
    \begin{minipage}{0.45\linewidth}
        \center
        \vspace{3mm}
        (a) PubMed article
    \end{minipage}
    \begin{minipage}{0.54\linewidth}
        \center
        \vspace{3mm}
        (b) MeSH hierarchy
    \end{minipage}
    \caption{Example of PubMed Article (a) and the partial MeSH hierarchy (b) that is associated with the article. For article (a), the PubMed database contains pmid, title, abstract, list of articles cited by, publication types, list of MeSH terms, and list of substances (chemicals). The MeSh hierarchy (b) shows the categorization of the MeSH terms to a broader concept.}
    \label{figure:pubmed}
\end{figure*}

Unlike existing heterogeneous bibliographic benchmarks, PubMed contains rich metadata beyond the citation structure.
Figure \ref{figure:pubmed}(a) shows an example of a PubMed article\footnote{Article can be found at \url{https://pubmed.ncbi.nlm.nih.gov/12429942/}.}.
In addition to the author, venue, and citation information that is commonly found in most bibliographic data, each PubMed article contains data regarding the Chemical Substances within the article, the type of article that characterizes the nature of the information or the type of research support received, and Medical Subject Headings (MeSH) terms which identify the broader concepts in the data. The categorical information of chemical substances and publication types are not found in DBLP, ACM, or MAG. Moreover, there are over 30,000 terms in the MeSH vocabulary, which exceeds the Computing Classification System (CCS) hierarchical ontology found in ACM. Furthermore, the terms follow a hierarchical taxonomy\footnote{\url{https://www.nlm.nih.gov/mesh/intro_trees.html}} (see Figure \ref{figure:pubmed}(b) for an example MeSH tree for some terms in the example), yet also have the unique property that a term can belong to one or more trees, unlike CCS.
Capturing this hierarchical structure can potentially improve the representation; however, the data in PubMed is incomplete as author disambiguation and hierarchical taxonomy are not available in the metadata.

We present \data, a new benchmark dataset of over 30 million PubMed articles for evaluating heterogeneous graph embeddings for biomedical literature.
It leverages the citations and author disambiguation capabilities of Semantic Scholar while also layering in the rich metadata that is offered in PubMed including the MeSH Terms, Chemical List, and Publication Type.
\data~also layers in the MeSH hierarchical structure for all the terms associated with the articles, which previous benchmarks do not support.
\data~provides 3 different tasks to evaluate the quality of the graph embeddings that span node classification, node clustering, and abstract screening for 21 SR tasks.
The latter task is different than the existing node-level and edge-level tasks provided in \neurips~and \kdd~in that the same node can have different labels depending on the SR content.
By providing a high-quality and large-scale heterogeneous bibliographic network with three different graph tasks and their associated evaluation metrics, we can measure progress in a consistent and reproducible fashion.

In addition to building \data, we perform extensive benchmark experiments for the dataset using current state-of-the-art graph embedding methods including 2 homogeneous GNNs and 3 heterogeneous graph embedding models.
Through the experiments, we highlight two research challenges associated with generating embeddings for PubMed.
First is the lack of scalability for many existing GNNs as many of the state-of-the-art models were incapable of processing the entire graph.
Second is the inability to capture the heterogeneity of the network as the models fail to achieve comparable performance to other bibliographic networks.
Our experimental results illustrate the need for developing new scalable heterogeneous GNN models that are capable of handling the rich metadata in PubMed, which is unlike any existing bibliographic data.
\section{Background}
Bibliographic data is used in various tasks, for example, word embedding using the title and abstract, network embedding using the citation, and author network.
Thus, many works have worked on constructing a benchmark for bibliographic data such as \neurips~\cite{hu2020open}, \kdd~\cite{lv2021we}, and \storc~\cite{lo2019s2orc}.
Here we briefly describe the three related academic paper benchmark datasets and their limitations.

\neurips~\cite{hu2020open} is a large-scale benchmark for graph machine learning tasks.
It encompasses a variety of domains such as social networks, biological networks, molecular graphs, and knowledge graphs.
\neurips~also has bibliographic data, for example, ogbn-arxiv and ogbn-papers100M are citation networks that are extracted from arxiv and MAG, respectively.
Notably, both of these are homogeneous networks with paper nodes and links that represent the citation.
\neurips~also has a heterogeneous academic network, ogbn-mag, which is extracted from MAG. 
The ogbn-mag dataset contains 4 different node types (i.e., papers, authors, institutions, and topics) along with their relations.
However, \neurips~focuses on benchmarking graph machine learning methods on the large-scale homogeneous network. 

\kdd~\cite{lv2021we} provided 11 medium-scale graph benchmark datasets for node classification, link prediction, and knowledge-aware recommendation.
For node classification, it contains DBLP, IMDB, ACM, and Freebase \cite{bollacker2008freebase} datasets and for link prediction, Amazon, LastFM, and PubMed datasets.
The PubMed benchmark is the subset of a previously generated network of genes, diseases, chemicals, and species filtered by domain experts \cite{yang2020heterogeneous}.
Furthermore, the PubMed dataset does not reflect the bibliographic data directly.
Instead, for \kdd, DBLP, and ACM datasets serve as the lone benchmarks for the bibliographic network. 
However, both datasets lack rich metadata that can be helpful for learning node embeddings.
Moreover, the benchmarks assume a single label for each node, whereas labels can change depending on the context.

\begin{table}[t]
    \small
    \centering
    \caption{Comparison of existing bibliographic datasets with N denoting nodes, NT denoting node types, ET denoting edge types and HIER denoting a hierarchical structure on at least one of the nodes.}
    \label{tab:comparison}
    \begin{tabular}{lrrrc}
        \toprule
                        & \# N          & \# NT & \# ET & HIER \\
        \midrule
        PGB             & 54,974,182    & 5     & 7     & \cmark \\
        ogbn-mag        & 1,939,743     & 4     & 4     & \xmark \\
        ogbn-arxiv      & 169,343       & 1     & 1     & \xmark \\
        ogbn-papers100M & 111,059,956   & 1     & 1     & \xmark \\
        HGB-DBLP        & 26,128        & 4     & 6     & \xmark \\
        HGB-ACM         & 10,942        & 4     & 6     & \xmark \\
        \bottomrule
    \end{tabular}
\end{table}

The \storc~\cite{lo2019s2orc} corpus is a large-scale academic paper corpus that is constructed using the data from the Semantic Scholar literature corpus \cite{ammar2018construction}.
Articles in Semantic Scholar are derived from numerous sources which are obtained directly from publishers such as MAG, arXiv, PubMed, and crawled from the open Internet. 
Semantic Scholar clusters these papers based on title similarity and DOI overlap, resulting in an initial set of approximately 200M paper clusters.
Using the Semantic Scholar literature corpus, S2ORC aggregated the metadata of articles and cleaned the data to select canonical metadata using external sources such as IEEE and DBLP.
Although \storc~contains biomedical literature, it mainly focuses on the common metadata that exists across all the articles. Since publication types, MeSH terms, and chemical substances are only present in biomedical literature, such metadata is not included in the dataset.
Thus, developing embeddings that reflect the heterogeneity of the PubMed database requires additional integration.
Table \ref{tab:comparison} summarizes the statistics of the existing bibliographic datasets.
\section{PubMed Graph Benchmark (\data)}
\begin{figure*}[t]
    \center
    \includegraphics[width=0.7\linewidth]{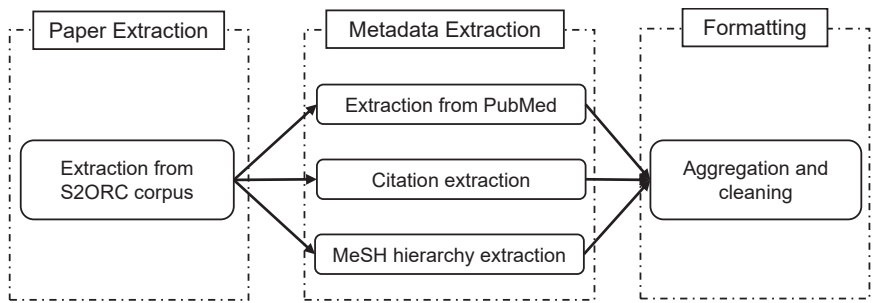}
    \caption{Framework overview of the \data.}
    \label{fig:overview}
\end{figure*}

In this section, we introduce the framework to construct \data, shown in Figure \ref{fig:overview}, and the format of the data, the evaluation tasks, and the license information.

\subsection{Paper Collection}
\data~is constructed based on the \storc~corpus \cite{lo2019s2orc} as it contains more complete citation information than PubMed.
However, there exist cases where the abstract only exists in the Semantic Scholar database but not in the PubMed database.
Since \data~targets the biomedical literature, we initially extract articles that contain a PubMed ID (PMID) from \storc.

\subsection{Metadata}
The \data~contains 13 metadata fields, which are shown in Table \ref{tab:metadata}. Here we detail the integrated fields from \storc~and various PubMed data sources.

\subsubsection{PubMed Extraction}
The \storc\ corpus only contains basic metadata of each PubMed paper (e.g., title, abstract, authors, year, and venue).
In biomedical literature, unlike general academic articles, there is important metadata that can serve an important role such as Medical Subject Headings (MeSH) terms and publication types.
Additional partial information can be found in PubMed (see Figure \ref{figure:pubmed}).
To extract more detailed information related to each article, we query information from the Entrez API\footnote{\url{https://www.ncbi.nlm.nih.gov/books/NBK25501/}} using the PMID.

The metadata contains ``Chemical List'', ``Publication Type'', and ``MeSH Terms".
The chemical list provides the registry number of specific chemical substances assigned by the Chemical Abstracts Service and the names of the chemical substances.
The publication type identifies the type of article indexed for MEDLINE and characterizes the nature of the information, how it is conveyed, and the type of research support received.
For example, an article can have a publication type of \emph{Review}, \emph{Letter}, \emph{Retracted Publication}, \emph{Research Support, N.I.H.}, or \emph{Clinical Conference}.
Finally, MeSH terms are used to characterize the content of the articles. 
MeSH terms are recorded with the information whether they are the major topic or not.
The major MeSH terms denote that those are the most significant topics of the paper whereas the non-major MeSH terms are used to identify concepts that have also been discussed in the item but are not the primary topics.
For the articles identified from our paper collection process, we integrate the names of the chemical substances, the publication type, and both major and minor MeSH terms.

\subsubsection{Citation Extraction}
While the PubMed database contains rich information on biomedical literature, it contains few information about the citations.
However, \storc~corpus extracted the citations from the collected PDF or LaTeX files on top of the Semantic Scholar literature corpus.
Thus, to construct \data, we first use the citation information from the \storc.
This includes both in and out citations which refer to whether the paper is cited by another paper or the paper cites another paper.
We convert all the Semantic Scholar IDs into PMIDs and remove papers that are not included in the PubMed database\footnote{While this can potentially harm or bias the embedding, we did this to maintain consistency in the article information in \data.}.
We note that there are cases where Semantic Scholar does not contain all the citations.
Thus we also extract citations from the Entrez API to include papers that do not exist in the Semantic Scholar database but exist in PubMed.
The metadata associated with the PMIDs of the newly identified papers is then retrieved to ensure consistency of the article information. 
In this fashion, the articles in \data~are not a pure subset of \storc.

\begin{table*}[t!]
    \centering
    \small
    \caption{List of metadata, the field name and the associated type for \data.}
    \label{tab:metadata}
    \begin{tabular}{llll}
    \toprule
    Field Description     & Field name        & Field type        & Explanation \\
    \midrule
    PMID                  & pmid                     & str        & Pubmed ID \\
    Title                 & title                    & str        & Title of the paper \\
    Abstract              & abstract                 & str        & Abstract of the paper \\
    Authors               & authors                  & List[Dict] & list of authors \\ 
    Year                  & year                     & int        & Published year \\
    Venue                 & venue                    & str        & Venue of the paper \\
    Publication Type      & publication\_type        & str        & Type of the article \\
    Chemical List         & chemicals                & List       & Name of the chemical substances \\
    MeSH Terms            & mesh                     & List[Dict] & List of MeSH terms \\
    Inbound Citation      & in\_citation             & List       & List of PMID that cites the paper \\
    Outbound Citation     & out\_citation            & List       & References of the paper \\
    Has Inbound Citation  & has\_inbound\_citations  & Boolean    & Validator for inbound citation \\
    Has Outbound Citation & has\_outbound\_citations & Boolean    & Validator for outbound citation \\
    \bottomrule
\end{tabular}
\end{table*}

\subsubsection{MeSH Terms Hierarchy}
One important feature of MeSH terms is the hierarchical ontology of the terms.
MeSH terms can be categorized into broader MeSH terms that support the categorization of the articles, as depicted in Figure \ref{figure:pubmed}(b).
The categories of different hierarchy levels reveal the similarity at coarser/fine-grained granularities.
As shown in Figure \ref{figure:pubmed}(b), MeSH terms that are assigned to the article can share the same parents or can be in a different sub-tree.
When comparing two articles, if they do not have the same MeSH terms but MeSH terms with the same parents (or within the same sub-tree), then the two articles are potentially closely related.
Therefore, knowing the hierarchy can play an important role in identifying similar articles.

Unfortunately, the Entrez API does not include the MeSH terms hierarchy. 
Thus, we also extract the MeSH terms hierarchy dataset\footnote{\url{https://www.nlm.nih.gov/databases/download/mesh.html}} to identify the position of the MeSH terms associated with each article.
The MeSH terms hierarchy dataset only contains the MeSH terms shown in Figure \ref{figure:pubmed}(b).
However, the tree numbers help reveal the hierarchical structure.
For example, the MeSH terms ``Chromosome Segregation" with tree number G04.144.220.220.625  and ``Mitosis" with tree number G04.144.220.220.781 demonstrate that they share the same parent, ``Cell Cycle" with tree number G04.144.220.220.
Thus, we integrate the tree number for each MeSH term using the MeSH terms hierarchy into \data.

\subsection{Data Format and Statistics}
Each article in \data~is stored using the JSON file format.
Table \ref{tab:metadata} summarizes the field name and the field type that are used to store \data.
The ``authors" field contains 4 subfields, ``first", ``middle", ``last", and ``suffix".
For the chemical list and the MeSH terms, we exclude the ids and only included the name because the name itself is already unique.
For the MeSH terms, we use 3 subfields to convey which MeSH terms are major or minor,
The subfield ``name" refers to the name of the MeSH terms, the subfield ``is\_major" is set to a true/false value to identify the major MeSH terms, and the subfield ``tree\_number" is the MeSH hierarchy information.
There can exist multiple major MeSH terms for each article.
We also included fields for validating whether the inbound and outbound citation exists in the benchmark.
The fields are named ``has\_outbound\_citations" and ``has\_inbound\_citations", and the value is set to be either true or false.
This helps users easily identify the presence of citation information without parsing the list of citations.

\begin{table}[t]
    \centering
    \caption{Statistics of \data~(30,872,730 articles).}
    \label{tab:stat}
    \begin{tabular}{l r r r}
        \toprule
        Name                           & Total \# & Missing (\%) & Avg per article \\
        \midrule
        Authors                        & 30,397,681 & 1.54  & 4.11 \\
        Articles w/ MeSH terms         & 26,883,163 & 12.92 & 9.32 \\
        Articles w/ chemical list      & 14,565,380 & 52.82 & 1.82 \\
        Articles w/ publication type   & 30,685,975 & 0.60  & 1.73 \\
        Articles w/ inbound citations  & 16,488,646 & 46.59 & 6.51 \\
        Articles w/ outbound citations & 7,781,767  & 74.79 & 6.51 \\
        \bottomrule
    \end{tabular}
\end{table}

The statistics of \data\ are shown in Table \ref{tab:stat}.
It contains 30,872,730 biomedical literature, and all the articles have PMID, title, abstract, year, venue, and at least one author.
However, there exist cases in which any one of the fields is missing which denote that the information does not exist in both PubMed and \storc.
46.59\% of articles do not have an inbound citation, and 74.49\% of articles do not have outbound citations.
Table \ref{tab:stat} also shows the average number of MeSH terms, chemicals, and inbound and outbound citations.
Due to the large size of the benchmark ($\sim$60GB), \data~is split into 10 partitions where each partition is compressed as a zip file.

\subsection{Evaluation Tasks}
The metadata in \data~contains all necessary information to construct a homogeneous or heterogeneous network.
There are 5 node types (\emph{P}aper, \emph{A}uthor, \emph{M}eSH terms, \emph{V}enue, and publication \emph{T}ype) and 7 edge types (P-P, P-A, A-A, P-M, P-V, P-T, M-M).
The constructed heterogeneous network can be used for node classification to determine the topic of articles, link prediction for citation recommendation, and SR for abstract/full-text screening.
We provide three evaluation tasks using \data.
Two of the tasks (node classification and clustering) follow a similar protocol to the tasks in existing benchmark datasets. The third task is new and specific to the biomedical nature of \data.

\subsubsection{Node Classification and Clustering}
We evaluated the network embedding methods on node classification and node clustering tasks.
These tasks are prevalent in existing graph benchmark datasets such as \neurips~and \kdd.
For both tasks, we use the labels provided by Namata \emph{et al.}~\cite{namata2012query} which consists of PubMed papers about diabetes.
Articles are labeled with 3 classes, `Diabetes Mellitus, Experimental', `Diabetes Mellitus Type 1', and `Diabetes Mellitus Type 2'.
The dataset encompasses 19,717 articles, 61,587 authors, and 4,081 MeSH terms.
However, the original dataset only contains 2 edge types, paper-author, and paper-MeSH terms.
We expand this dataset to include one more node (publication type) and two edge types, paper-paper, and paper-publication type.

\textit{Evaluation Metrics}. For the node classification task, we adopt micro and macro F1-score as the evaluation metrics as it is a multi-classification task (i.e., 3 classes). To assess the quality of the clusters, we use normalized mutual information (NMI) and adjusted rand index (ARI). For the number of clusters, we follow the number of classes used for the node classification task. The supplemental material contains detailed information regarding the calculation of these metrics. 

\begin{figure*}[t!]
    \centering
    \includegraphics[width=0.88\linewidth]{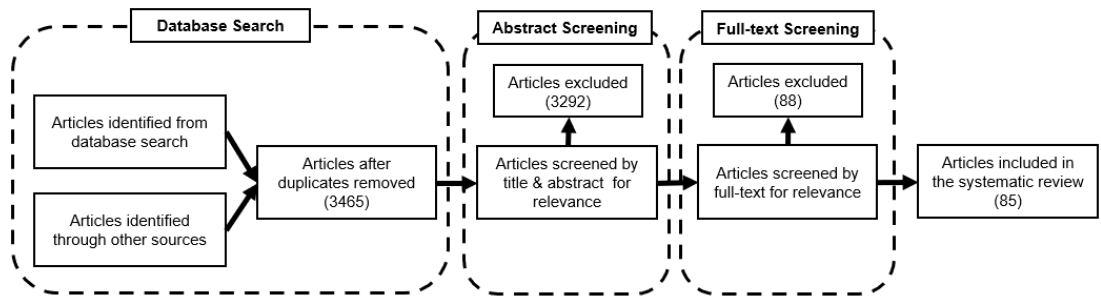}
    \caption{A simplified illustration of the SR screening process using ``Statins'' from Cohen dataset \cite{cohen2006reducing}.}
    \label{fig:sr}
\end{figure*}

\subsubsection{Systematic Review}
Systematic reviews (SRs) are essential knowledge translation tools focused on bridging the research-to-practice gap across a wide range of domains.
In health research, SRs aims to identify, evaluate, and summarize the findings of all individual studies (which typically describe clinical trial results) relevant to a clinical question, thereby making the available evidence more accessible \cite{Chalmers:2016it, Gough:2002fd,gough2017introduction}.
SRs serve as the basis for evidence-based medicine. 

Figure \ref{fig:sr} shows an illustration of the SR screening process.
The first step is to retrieve the initial list of articles using the combination of keywords and MeSH terms related to the topic (3465 articles from the figure).
Once the initial list is retrieved, reviewers go over the title and abstract screening to get rid of irrelevant articles (95\% of the articles).
Finally, the reviewers go over the full-text for articles passing the title and abstract screening phase to collect the relevant articles to the topic (2.45\% of articles).
However, the broad searches yield imprecise search results (e.g., <2\% relevant documents) and result in a labor-intensive process.
Current estimates for conducting an SR is 67 weeks from registration to publication \cite{borah2017analysis}.
Clearly, this process is neither unsustainable nor scalable, especially given the exponential growth of biomedical literature \cite{Bastian:2010hy}.
We currently integrate three different SR datasets: the popular, and publicly available dataset provided by Cohen \emph{et al}. \cite{cohen2006reducing}, SWIFT-Review \cite{howard2016swift} and CLEFT-TAR \cite{kanoulas2019clef}.

The Cohen \emph{et al}. dataset\footnote{\url{https://dmice.ohsu.edu/cohenaa/systematic-drug-class-review-data.html}} was the first SR dataset publicly released and was based on comparing 15 classes of drugs to treat specific conditions \cite{cohen2006reducing}. The evidence reports were completed by three evidence-based practice centers in Oregon, Southern California, and the Research Triangle Institute / University of North Carolina. 
The staff at these centers created search queries to identify randomized controlled trials by combining the health conditions and interventions.
Another set of datasets is 3 sets provided by SWIFT-Review \cite{howard2016swift}.\footnote{\url{https://systematicreviewsjournal.biomedcentral.com/articles/10.1186/s13643-016-0263-z\#Sec30}}
The dataset was generated by the National Toxicology Program (NTP) Office of Health Assessment and Translation (OHAT).
The last set is provided by CLEF 2019 e-Health TAR Lab \cite{kanoulas2019clef} (Task 2)\footnote{\url{https://github.com/CLEF-TAR/tar}} which focuses on retrieving relevant studies from during the abstract and title screening phase of conducting an SR.
From the CLEF-TAR dataset, we randomly selected 3 sets which are the CD012661 topic from Prognosis, the CD008803 topic from DTA, and the CD005139 topic from Intervention.
The title of the topic CD012661 is ``Development of type 2 diabetes mellitus in people with intermediate hyperglycemia'' \cite{richter2018development}.
The title of the topic CD008803 is ``Optic nerve head and fibre layer imaging for diagnosing glaucoma"\cite{michelessi2015optic}.
The title of the topic CD005139 is ``Anti‐vascular endothelial growth factor for neovascular age‐related macular degeneration" \cite{solomon2014anti}.

Each article, identified using the PMID, was triaged using a two-step process.
First, the abstract is reviewed to determine if it meets the inclusion criteria of the SR.
If the criteria are met, the full text of the article is then reviewed to determine if the evidence should be summarized in the SR.
\data~targets the abstract screening process where most of the articles are excluded.
Table \ref{tab:sr_dataset} summarizes the statistics for each SR topic.
The abstract (abs) denotes whether the article passed the title/abstract screening phase.
It is notable that the number of articles included after the abstract screening varies from 1.44\% to 32.48\%, demonstrating a relatively large degree of imbalance across the SR topics.

\begin{table}[t]
\small
    \centering
    \caption{Statistics of all datasets used. Abs refers to the number of articles passing the abstract triage statuses and \% shows the percentage.}
    \label{tab:sr_dataset}
    
    \begin{tabular}{lrrrr}
        \toprule
        SR                      & Abs & Total & \%    \\
        \midrule
        Cohen-ACEInhibitors           & 183 & 2544  & 7.23  \\
        Cohen-ADHD                    & 84  & 851   & 9.87  \\
        Cohen-Antihistamines          & 92  & 310   & 29.67  \\
        Cohen-AtypicalAntipsychotics  & 363 & 1120  & 32.41 \\
        Cohen-BetaBlockers            & 302 & 2072  & 14.57  \\
        Cohen-CalciumChannelBlocker   & 279 & 1218  & 22.90  \\
        Cohen-Estrogens               & 80  & 368   & 21.74 \\
        Cohen-NSAIDs                  & 88  & 393   & 22.39 \\
        Cohen-Opioids                 & 48  & 1915  & 2.51  \\
        Cohen-OralHypoglycemics       & 139 & 503   & 27.63 \\
        Cohen-ProtonPumpInhibitors    & 238 & 1333  & 17.85  \\
        Cohen-SkeletalMuscleRelaxants & 34  & 1643  & 2.56  \\
        Cohen-Statins                 & 173 & 3465  & 4.99  \\
        Cohen-Triptans                & 218 & 671   & 32.48  \\
        Cohen-UrinaryIncontinence     & 78  & 327   & 23.85 \\
        \midrule
        SWIFT-Transgenerational       & 765 & 48638 & 1.57 \\
        SWIFT-PFOS-PFOA               & 95  & 6331  & 1.50 \\
        SWIFT-BPA                     & 111 & 7700  & 1.44 \\
        \midrule
        CLEF-Prognosis-CD012661       & 192 & 3367  & 5.70 \\
        CLEF-DTA-CD008803             & 99  & 5220  & 1.89 \\
        CLEF-Intervention-CD005139    & 112 & 5392  & 2.07 \\
        \bottomrule
    \end{tabular}
\end{table}

\textit{Evaluation Metrics}. Cohen \emph{et al}.~\cite{cohen2006reducing} introduced a measure \emph{work saved over sampling} (WSS) which measures the work saved over random sampling for a given level of recall.
Several works also evaluated using the area under the receiver operating curve (AUC) for predicting whether or not the abstract was screened or not to report the results \cite{cohen2008optimizing,miwa2014reducing}.
For the purpose of this report, we report both AUC and WSS scores.

\subsection{Code and Data License Information}
The entire data is released and publicly available on Zenodo.\footnote{\url{https://zenodo.org/record/6406776\#.YqrOKnbMKUk}}
We open-source the code to reconstruct the benchmark in a GitHub repository.\footnote{\url{https://github.com/ewhlee/PGB}}.
The GitHub repository also contains the train, validation, and test splits for the three evaluation tasks.
We hope that by releasing the code publicly, the community can contribute to the maintenance of the benchmark dataset (i.e., updating the graph or adding new tasks).

\data~is released under the CC BY-NC 4.0 license and for non-commercial use. \data~is constructed using the PubMed Entrez API and \storc. \storc~is non-commercial use and released under the same license (CC BY-NC 4.0 license). The PubMed Entrez API does not require a signed license agreement to download publicly accessible data. However, we note that the associated PubMed metadata (i.e., MeSH terms, Chemical list, and publication type) in \data~may not reflect the most current data available on PubMed. The data can be re-updated using the Github repository assuming no major changes in the type and format of the machine-readable data. The usage guidelines and registration for the API key are detailed in the electronic book chapter\footnote{\url{https://www.ncbi.nlm.nih.gov/books/NBK25497/\#chapter2.Usage\_Guidelines\_and\_Requiremen}}. Note that potential publication bias or other ethical considerations may need to be considered further.

\section{Experiments}
In this section, we discuss the experimental settings for the evaluation tasks.

\subsection{Experimental Design}
For the Systematic Review task, we construct 3 different subsets of \data~for computational reasons.
We trace the inbound and outbound citations up to 2-hops from the original articles to construct 3 subnetworks of approximately 1.2M, 3.4M, and 1.8M articles, respectively for the Cohen, SWIFT, and CLEF-TAR datasets.
In each subnetwork, we randomly split the graph into train-validation-test by sampling articles within each SR task using a 50\%-25\%-25\% ratio. We create 3 train-validation-test trials for each subnetwork.
For all the baselines, we used a g4dn AWS instance with NVIDIA T4 GPU.

\subsection{Baseline Models and Hyperparameters}
We evaluate various 6 different models that include document embedding, homogeneous network embedding, heterogeneous embedding, and knowledge graph embedding models.
\begin{itemize}
    \item SPECTER \cite{specter2020cohan}: The SPECTER is an embedding model that learns the representation of a document by computing the embeddings using a SciBERT model \cite{beltagy2019scibert} pre-trained on relatedness signals derived from the citation graph. We use the embeddings for SPECTER provided by Semantic Scholar API.\footnote{\url{https://www.semanticscholar.org/product/api}} The Semantic Scholar API allows a paper search by using the PubMed ID to retrieve the Semantic Scholar ID. Using this Semantic Scholar ID, we retrieve the SPECTER embeddings of each document.
    \item LINE \cite{tang2015line}: LINE is a conventional homogeneous network embedding method that uses first- and second-proximity. 
    LINE uses the joint probability between two nodes.\footnote{\url{https://github.com/DeepGraphLearning/graphvite}.} We set the number of dimensions to 128 for both first- and second-proximity. The final embedding is the concatenation of 2 proximities. As LINE is an unsupervised model, we add a soft-max layer on top of the final embeddings.
    \item GCN \cite{kipf2016semi}: GCN is a graph convolutional network embedding model designed for a homogeneous network.\footnote{\url{https://github.com/tkipf/gcn}.} GCN is trained in a supervised setting using the SR task. We use the 500-dimension TF-IDF weighted word vector provided by Namata \emph{et al.}~\cite{namata2012query} as the node feature.
    \item HAKE \cite{zhang2020learning}: HAKE is a hierarchical-aware knowledge graph embedding model which is not a GNN-based model but a translational distance model which describes relations as translations from one node to the other.\footnote{\url{https://github.com/MIRALab-USTC/KGE-HAKE}.} It uses radial coordinates to embed entities at different levels of the hierarchy and uses angular coordinates to distinguish entities at the same level of the hierarchy.  
    HAKE uses the link-prediction task to learn the embeddings, and thus it is an unsupervised model. For the supervised tasks, we add a soft-max layer on top of the embeddings. We try [500, 1000] for the dimension size and select 1000 as the validated parameter.
    \item GAHNE \cite{li2020gahne}: GAHNE is a model to learn representations for HIN.\footnote{\url{https:github.com/seanlxh/GAHNE}.} It converts the network into a series of homogeneous sub-networks to capture semantic information. An aggregation mechanism then fuses the sub-networks with supplemental information from the whole network. Using the validation set, we process a grid search using [0.01, 0.005, 0.001] for the learning rate, [0.0005, 0.001] for the L2 penalty, and [64, 128, 256] for the dimension size. The validated parameters we used are a learning rate of 0.005, a dropout of 0.5, an L2 penalty of 0.001, and a dimension of 128. GAHNE is a supervised model and the model is trained using the labels from SR topics.
    \item ie-HGCN \cite{yang2020interpretable}: ie-HGCN is a GCN-based HIN embedding model that evaluates all possible meta-paths and projects the representations of different types of neighbor objects into a common semantic space using object- and type-level aggregation.\footnote{\url{https://github.com/kepsail/ie-HGCN/}.} We use the supervised, cross-entropy loss to learn the weights of the model. We set the number of layers to be 5 and using the validation set, we tried [(128, 64, 32, 16), (156, 128, 64, 32)] as the dimension size. The validated parameters used in the results are 5 layers, with the dimensions of input, 128, 64, 32, and 16. We use the same node feature as GCN.
\end{itemize}

LINE and HAKE are unsupervised models, and the other 3 GNN baselines are semi-supervised models.
For the Systematic Review task, for the homogeneous network, we only use the citation information to construct the network.
For HAKE, we use 3 node types (\emph{P}aper, \emph{M}eSH terms, publication \emph{T}ype) with 4 edge types (P-P, P-M, M-M, P-T). For the other 2 heterogeneous networks, we use 4 node types (\emph{P}aper, \emph{V}enue, \emph{M}eSH terms, publication \emph{T}ype) with 4 edge types (P-P, P-V, P-M, and P-T)

We use the code and perform a parameter search around the neighborhood of suggested parameters provided by the original paper. We briefly describe the final parameters used and provide a link to the implementation. For each of the implementations, we kept separate environment files to ensure that the required Python packages were installed and the correct version as outlined in the code.
The validation set is used to tune the hyperparameters for GAHNE and ie-HGCN.
\begin{table}[t]
    \centering
    \caption{Network embedding result of 3 trials. The best score for each node classification and clustering is bolded and the second highest is underlined.}
    \label{tab:class}
    \begin{tabular}{lcccc}
    \toprule
    Baseline  & Macro-F1 & Micro-F1  & NMI   & ARI \\
    \midrule
    LINE    & 36.47          & 39.67          & 6.56           & 5.38           \\
    GCN     & 42.13          & 43.19          & 6.98           & 5.96           \\
    HAKE    & 48.71          & 50.85          & 10.73          & 10.11          \\
    GAHNE   & \textbf{50.44} & \textbf{53.54} & \textbf{13.86} & \textbf{13.52} \\
    ie-HGCN & \underline{50.37}    & \underline{53.37}    & \underline{13.71}    & \underline{13.27}    \\
    \bottomrule
\end{tabular}
\end{table}

\section{Evaluations}
In this section, we discuss the performance of the various models using the subset of \data.

\subsection{Node Classification and Clustering}

The results of the performance for node classification (macro and micro F1-score) and clustering (NMI and ARI) are shown in Table \ref{tab:class} with the best results bolded and the second best results underlined.
The heterogeneous network embedding models (HAKE, GAHNE, and ie-HGCN) outperform the homogeneous network embedding models (LINE and GCN), illustrating that modeling the multiple node types and link types is beneficial.
Both GAHNE and ie-HGCN have similar scores across all four metrics.
The difference between LINE and GCN shows the importance of using the word information as GCN uses the TF-IDF weighted word vectors for the node feature on top of the citation network while LINE only uses the citation network.
Unfortunately, a major limitation of existing heterogeneous network embedding models is the memory footprint.
We tried other heterogeneous network embedding models such as GTN \cite{yun2019graph}, HetGNN \cite{zhang2019heterogeneous}, and MAGNN \cite{fu2020magnn}, but the models ran out of memory.

\subsection{Systematic Review}
All the results shown in this section use the subnetwork of each dataset (Cohen, SWIFT-Review, and CLEF-TAR).
We compare the performance of 1 language model and 5 network embedding models on SR.
The performance is reported in Table \ref{tab:sr_auc} in the average of AUC scores of 3 trials for each SR task and in Table \ref{tab:sr_wss} in the average of WSS scores with the same setting.
The best results are bolded and the second-best results are underlined.

\begin{table*}[t]
    \centering
    \caption{SR statistics and average AUC results across the 3 trials for the various models. The best score is bolded and the second highest is underlined.}
    \label{tab:sr_auc}
    \footnotesize
    \begin{tabular}{lcccccc}
        \toprule
                SR Topic                      & SPECTER & LINE  & GCN   & HAKE  & GAHNE          & ie-HGCN \\ 
        \midrule
                Cohen-ACEInhibitors           & 0.677   & 0.580 & 0.592 & 0.677 & \underline{0.731} & \textbf{0.740}    \\
        Cohen-ADHD                    & 0.567   & 0.548 & 0.577 & 0.599 & \underline{0.600} & \textbf{0.607}    \\
        Cohen-Antihistamines          & 0.505   & 0.493 & 0.509 & 0.521 & \textbf{0.558}    & \underline{0.542} \\
        Cohen-AtypicalAntipsychotics  & 0.638   & 0.555 & 0.597 & 0.648 & \textbf{0.708}    & \underline{0.699} \\
        Cohen-BetaBlockers            & 0.699   & 0.586 & 0.606 & 0.683 & \textbf{0.733}    & \underline{0.728} \\
        Cohen-CalciumChannelBlockers  & 0.601   & 0.594 & 0.608 & 0.621 & \textbf{0.654}    & \underline{0.651} \\
        Cohen-Estrogens               & 0.637   & 0.544 & 0.588 & 0.647 & \textbf{0.676}    & \underline{0.673} \\
        Cohen-NSAIDS                  & 0.694   & 0.586 & 0.615 & 0.690 & \textbf{0.767}    & \underline{0.746} \\
        Cohen-Opiods                  & 0.675   & 0.603 & 0.637 & 0.686 & \underline{0.725} & \textbf{0.727}    \\
        Cohen-OralHypoglycemics       & 0.535   & 0.512 & 0.529 & 0.533 & \textbf{0.567}    & \underline{0.557} \\
        Cohen-ProtonPumpInhibitors    & 0.674   & 0.604 & 0.626 & 0.681 & \textbf{0.731}    & \underline{0.729} \\
        Cohen-SkeletalMuscleRelaxants & 0.688   & 0.605 & 0.632 & 0.687 & \underline{0.724} & \textbf{0.733}    \\
        Cohen-Statins                 & 0.668   & 0.572 & 0.608 & 0.662 & \underline{0.710} & \textbf{0.716}    \\
        Cohen-Triptans                & 0.658   & 0.587 & 0.618 & 0.668 & \textbf{0.723}    & \underline{0.717} \\
        Cohen-UrinaryIncontinence     & 0.696   & 0.605 & 0.633 & 0.681 & \textbf{0.745}    & \underline{0.741} \\
        \midrule
        SWIFT-Transgenerational       & 0.695   & 0.637 & 0.667 & 0.684 & \underline{0.741} & \textbf{0.761}    \\
        SWIFT-PFOS-PFOA               & 0.671   & 0.634 & 0.657 & 0.695 & \underline{0.721} & \textbf{0.728}    \\
        SWIFT-BPA                     & 0.632   & 0.563 & 0.604 & 0.645 & \underline{0.725} & \textbf{0.729}    \\
        \midrule
        CLEF-Prognosis-CD012661       & 0.678   & 0.593 & 0.628 & 0.647 & \underline{0.671} & \textbf{0.691}    \\
        CLEF-DTA-CD008803             & 0.619   & 0.598 & 0.628 & 0.643 & \underline{0.681} & \textbf{0.691}    \\
        CLEF-Intervention-CD005139    & 0.665   & 0.623 & 0.646 & 0.666 & \underline{0.702} & \textbf{0.704}    \\
        \bottomrule
    \end{tabular}
\end{table*}

\begin{table*}[t]
    \centering
    \caption{SR statistics and average WSS results across the 3 trials for the various models. The best score is bolded and the second highest is underlined.}
    \label{tab:sr_wss}
    \footnotesize
    \begin{tabular}{lcccccc}
\toprule
SRTopic                       & SPECTER & LINE  & GCN   & HAKE        & GAHNE          & ie-HGCN        \\
\midrule
Cohen-ACEInhibitors           & 0.388   & 0.343 & 0.364 & 0.385       & \underline{0.472}    & \textbf{0.489} \\
Cohen-ADHD                    & 0.274   & 0.247 & 0.253 & 0.277       & \underline{0.343}    & \textbf{0.344} \\
Cohen-Antihistamines          & 0.111   & 0.042 & 0.079 & 0.109       & \textbf{0.168} & \underline{0.137}    \\
Cohen-AtypicalAntipsychotics  & 0.092   & 0.059 & 0.066 & 0.087       & \textbf{0.111} & \underline{0.102}    \\
Cohen-BetaBlockers            & 0.209   & 0.186 & 0.19  & 0.211       & \underline{0.291}    & \textbf{0.304} \\
Cohen-CalciumChannelBlockers  & 0.21    & 0.173 & 0.194 & 0.208       & \underline{0.221}    & \textbf{0.242} \\
Cohen-Estrogens               & 0.223   & 0.169 & 0.197 & 0.222       & \textbf{0.259} & \underline{0.256}    \\
Cohen-NSAIDS                  & 0.385   & 0.377 & 0.384 & 0.437       & \textbf{0.508} & \underline{0.505}    \\
Cohen-Opiods                  & 0.253   & 0.21  & 0.218 & 0.276       & \underline{0.339}    & \textbf{0.343} \\
Cohen-OralHypoglycemics       & 0.111   & 0.057 & 0.065 & 0.102       & \textbf{0.133} & \underline{0.128}    \\
Cohen-ProtonPumpInhibitors    & 0.233   & 0.194 & 0.204 & 0.249       & \textbf{0.287} & \underline{0.283}    \\
Cohen-SkeletalMuscleRelaxants & 0.198   & 0.143 & 0.165 & 0.204       & \underline{0.239}    & \textbf{0.246} \\
Cohen-Statins                 & 0.229   & 0.169 & 0.179 & 0.227       & \underline{0.255}    & \textbf{0.256} \\
Cohen-Triptans                & 0.343   & 0.278 & 0.294 & 0.348       & \textbf{0.372} & \underline{0.362}    \\
Cohen-UrinaryIncontinence     & 0.21    & 0.162 & 0.174 & 0.202       & \textbf{0.233} & \underline{0.232}    \\
\midrule
SWIFT-Transgenerational       & 0.202   & 0.111 & 0.155 & 0.191       & \underline{0.253}    & \textbf{0.277} \\
SWIFT-PFOS-PFOA               & 0.241   & 0.195 & 0.203 & 0.258       & \underline{0.378}    & \textbf{0.383} \\
SWIFT-BPA                     & 0.354   & 0.258 & 0.287 & \underline{0.376} & \textbf{0.441} & \textbf{0.441} \\
\midrule
CLEF-Prognosis-CD012661       & 0.207   & 0.152 & 0.164 & 0.205       & \textbf{0.252} & \underline{0.248}    \\
CLEF-DTA-CD008803             & 0.302   & 0.219 & 0.222 & 0.297       & \textbf{0.341} & \underline{0.337}    \\
CLEF-Intervention-CD005139    & 0.2     & 0.143 & 0.158 & 0.199       & \underline{0.278}    & \textbf{0.283} \\
\bottomrule
\end{tabular}
\end{table*}

As shown in the tables, both of the results (AUC and WSS scores) of the heterogeneous network embedding models (HAKE, GAHNE, and ie-HGCN) significantly outperform the homogeneous network embedding models (LINE and GCN).
This suggests that not only the citation information but also other node types (venue, MeSH terms, and publication type) help to improve the performance of the systematic review task.
GAHNE and ie-HGCN outperform HAKE as HAKE is an unsupervised model while others are semi-supervised models.
However, by comparing the performance with the homogeneous model, HAKE shows the importance of the hierarchical information (MeSH hierarchy).
The performance between GAHNE and ie-HGCN is similar.
The results suggest that ie-HGCN performs better when there are more articles excluded from the abstract screening phase.
For example, the ``SWIFT-BPA" dataset has a total of 7700 articles in the beginning but only 111 articles (1.44\%) are selected.
Whereas ie-HGCN performs better in cases when fewer articles are selected, GAHNE performs better in cases when more papers are selected.
For example, ``Cohen-AtypicalAntipsychotics'' starts with 1120 articles, and 363 articles (32\%) passed the screening. 

By comparing with the language model (SPECTER), it shows similar results with HAKE.
In other words, SPECTER outperforms the homogeneous network embedding models (LINE and GCN) which only uses the citation network but underperforms the heterogeneous network embedding models (GAHNE and ie-HGCN).
Although SPECTER is based on the transformer language model, it uses the document-level relatedness from the citation graph.
Thus, this helps SPECTER to outperform the supervised homogeneous network embedding models.
This illustrates the importance of both the abstract and the citation graph in the systematic review process.
Yet, even integrating the text and citation together does not beat the rich contextual information found in the venue, MeSH terms, and publication type.
\section{Discussions and Limitations}
Besides the baselines which are all embedding models, we also use a majority class classifier for the SR task.
The AUC score for using the majority class classifier is 0.5 for all SR tasks.
However, the accuracy is different for each SR task. For example, ``ADHD'' is 0.891, and ``Statins" is 0.95.
The AUC score is less sensitive to class imbalance as the minority class will have a strong impact on the AUC score, which illustrates the difficulty of predicting class imbalance datasets.
On the other hand, accuracy is not sensitive to class imbalance, thus it can be very high as our result even the minority class is not well predicted.

We note that our work has several limitations that could be improved upon. First, citations that were not a part of PubMed were not included. This was done because there would have been insufficient and inconsistent metadata with the rest of the articles in PGB, as well as the need to find some identifier for the articles themselves to denote that they were distinct from the PubMed articles. Second, many of the existing baseline methods were unable to scale to the entire dataset. Such models typically require considerable computing resources which we, unfortunately, did not have access to. More extensive methods can be included for future work to better understand the performance of state-of-the-art models. Finally, we note that the SR dataset is based on old evidence reports, while there has been a considerable number of recent SRs on various topics. We are in the process of expanding this to include more SRs by collaborating with people who have conducted recent SRs in public health, nursing, and medicine.
\section{Conclusion}
In this paper, we discuss the importance of biomedical literature and the necessary metadata fields for research.
There exist many studies that use heterogeneous network embedding for various tasks such as node classification, link prediction, and SR.
However, it is time-consuming to aggregate the necessary information from multiple sources to capture rich bibliographic data.
We construct \data, a biomedical literature bibliographic dataset, that contains 11 fields of metadata.
The strength of \data~is not only that it contains multiple types of nodes and edges, but also captures a hierarchical structure on one of its nodes.
Our experimental results illustrate that the scalability and the capability of handling rich metadata, especially the hierarchical structure, for existing graph embedding models, still remain open challenges.

\bibliographystyle{plain}
\bibliography{references}

\end{document}